\documentclass{svproc}
\usepackage{url}

\usepackage{graphicx} 
\usepackage{amssymb}
\usepackage{algorithm}
\usepackage{algorithmic}
\usepackage{multirow}
\usepackage{bm}
\usepackage{caption} 
\usepackage{bbding}
\def\V#1{\boldsymbol{\mathit#1}}
\usepackage{newfloat}
\begin{document}
\mainmatter              
\title{Q-Net: Query-Informed Few-shot Medical Image Segmentation}
\titlerunning{Q-Net: Query-Informed Few-shot Medical Image Segmentation}  
%
\author{Qianqian Shen \and Yanan Li$^($\Envelope $^)$ \and Jiyong Jin \and Bin Liu$^($\Envelope $^)$ \thanks{Code is available at https://github.com/ZJLAB-AMMI/Q-Net}} 
%
\authorrunning{Shen et al.} 
%
\tocauthor{Qianqian Shen, Yanan Li, Jiyong Jin, and Bin Liu}
\institute{Research Center for Applied Mathematics and Machine Intelligence,\\ Zhejiang Lab, Hangzhou 311121, China,\\
\email{\{shenqq,liyn,jinjy,liubin\}@zhejianglab.com}}

\maketitle              

\begin{abstract}
Deep learning has achieved tremendous success in computer vision, while medical image segmentation (MIS) remains a challenge, due to the scarcity of data annotations. Meta-learning techniques for few-shot segmentation (Meta-FSS) have been widely used to tackle this challenge, while they neglect possible distribution shifts between the query image and the support set. In contrast, an experienced clinician can perceive and address such shifts by borrowing information from the query image, then fine-tune or calibrate her prior cognitive model accordingly. Inspired by this, we propose Q-Net, a Query-informed Meta-FSS approach, which mimics in spirit the learning mechanism of an expert clinician. We build Q-Net based on ADNet, a recently proposed anomaly detection-inspired method. Specifically, we add two query-informed computation modules into ADNet, namely a query-informed threshold adaptation module and a query-informed prototype refinement module. Combining them with a dual-path extension of the feature extraction module, Q-Net achieves state-of-the-art performance on widely used abdominal and cardiac magnetic resonance (MR) image datasets. Our work sheds light on a novel way to improve Meta-FSS techniques by leveraging query information.
\end{abstract}
\section{Introduction}
\begin{figure}[t!]
	\centering
	\includegraphics[width=0.8\textwidth,scale=1.3]{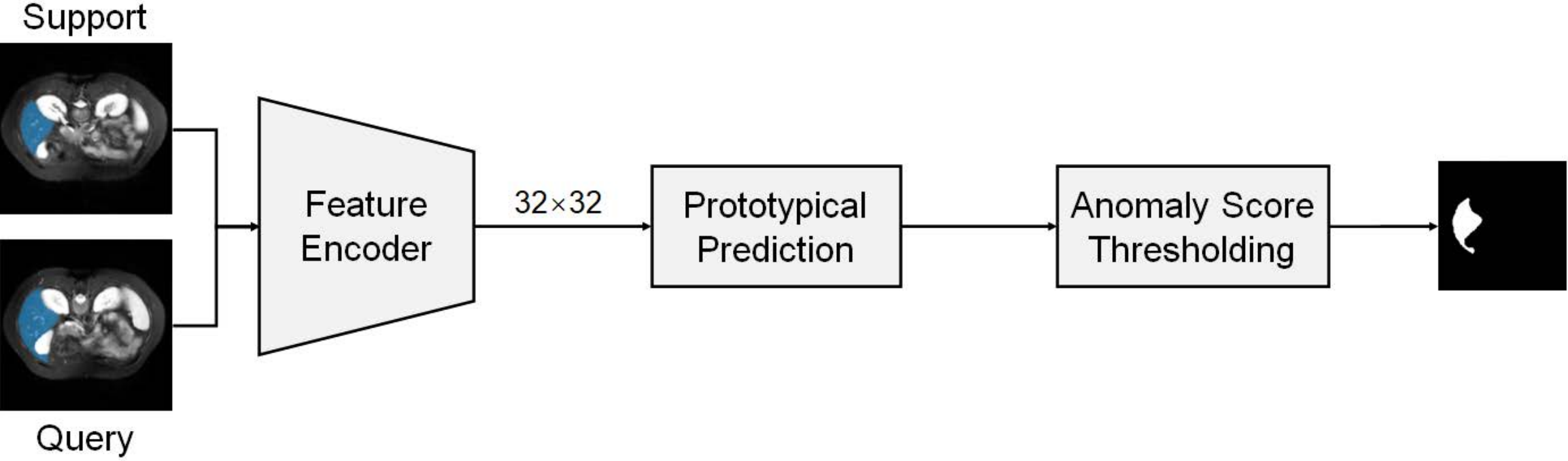}
	\centerline{(a) The prior art ADNet~\cite{hansen2022anomaly}}
	\hfill
	\includegraphics[width=0.8\textwidth,scale=1.2]{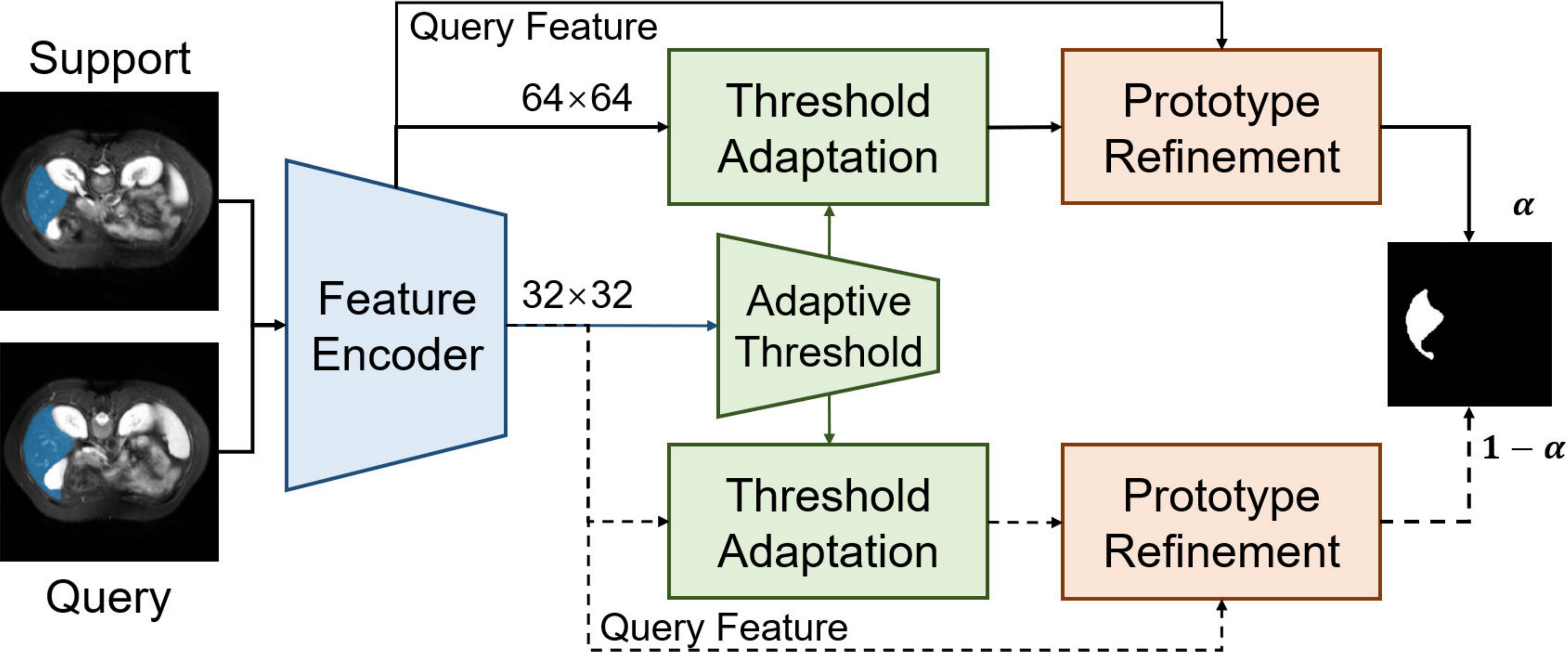}
	\centerline{(b) The proposed Q-Net}
	\caption{\small Comparison between (a) ADNet~\cite{hansen2022anomaly} and (b) our proposed Q-Net during inference. Q-Net differs from ADNet in 3 ways. First, it performs a query-informed threshold adaptation, yielding an adaptive, rather than fixed, threshold for use in anomaly-detection. Second, it has an additional query-informed prototype refinement module. Lastly, it runs in dual-path, capturing dual-scale features. The feature extractor and adaptive threshold generator is determined during meta-training. During inference, a query-informed prototype refinement module refines the initial class prototype to improve the final segmentation.}
	\label{fig1}
\end{figure}

Segmentation of organs, tissues, and lesions from biomedical images plays a key role in clinical research, including disease diagnosis, treatment planning and rehearsal, and surgical guidance~\cite{wang2018deepigeos,hesamian2019deep}. Convolutional neural networks (CNNs) have achieved state-of-the-art (SOTA) performance in semantic segmentation for both natural~\cite{noh2015learning,long2015fully,chen2018encoder} and medical images~\cite{ronneberger2015,milletari2016,wang2021automatic}. The widely used segmentation frameworks need a large number of labeled images for model training. Such a supervised learning paradigm is not practical for medical scenarios, as annotating medical images requires clinical expertise, which is much more expensive than labeling natural images. There are some publicly available biomedical datasets, which have annotations for major human organs among the liver, kidney and spleen, while annotations for other organs such as adrenal glands and duodenum are scarce and hard to collect~\cite{orting2019survey,kavur2021chaos,sun2022few}.

A natural strategy for machine learning using scarce data annotations is model fine-tuning. However, it does not work well for MIS, since adapting a pre-trained model to segment objects of a novel class based on extremely few labeled images can easily render over-fitting. Recently, meta-learning techniques for FSS have been widely used to tackle the above challenge~\cite{shaban2017one,rakelly2018few,wang2019panet,zhang2020sg}. In particular, they learn to segment objects of data-scarce classes over many simulated segmentation tasks in an episodic fashion. Such simulated tasks are designed with use of images of data-rich classes, and each task is assigned with a support set and a query set, mimicking a test-time scenario corresponding to a specific type of organ or lesion. The support set is used to learn discriminative representations of each class, which are then used to do segmentation predictions on the query set~\cite{dong2018few}. After meta-training over these tasks, the resulting algorithm agent is hoped to be capable of  segmenting a previously ``unseen" organ or lesion based solely on a few annotated images.

In contrast with natural images, medical images have their own characteristics that bring new challenges. First, the imbalance between the foreground object and the background is more severe than for natural images. For medical images, the texture of the foreground object is generally homogeneous, while the background is usually spatially in-homogeneous due to the existence of abundant tissues and objects of ``unseen" classes~\cite{sun2022few}. Further, human organs have large variations in size, shape, and location across different patients. A specific organ or tissue may have different appearance patterns due to different image acquisition protocols and different imaging modalities given by different equipments. Even in the same 3D volume, the 2D slices related to the same organ can be dissimilar.

The aforementioned issues can lead to distribution shifts between the query image and the support set, while, to our knowledge, current techniques like prototypical networks, do not take into account such shifts explicitely. In clinical practice, an experienced clinician can borrow information from the query image, such as the CT or MRI scans of a patient, to fine-tune or calibrate her prior knowledge for determining the size, shape and location of a target tissue. In this way, the distribution shift is removed or at least mitigated.
Inspired by the above observation, we propose Q-Net, a query-informed FSS approach that mimics in spirit the learning mechanism of an expert clinician. Figure~\ref{fig1} shows Q-Net compared to the prototypical network method based on which it is developed.

\textbf{Our Contributions}:
\begin{itemize}
\item[-] Inspired by the learning mechanism of expert clinicians mentioned above, we put forward a query-informed calibration (QIC) strategy to improve prior art meta-FSS techniques. As an instantiation, we show how to improve a recent SOTA network, namely ADNet ~\cite{hansen2022anomaly}, by introducing QIC computational modules, as shown in Figure ~\ref{fig1}. The resulting network is coined Q-Net, whose architecture is shown in Figure ~\ref{fig2}. Unlike ADNet that uses a fixed threshold and a static prototype during inference, Q-Net can adjust its threshold and prototype automatically to mitigate possible distribution shifts between the query image and the support set. Besides, Q-Net uses a dual-path architecture to capture double-scale features.
\item[-] We validate the performance of Q-Net experimentally on two widely-used medical image datasets. Results show that our Q-Net significantly outperforms other SOTA methods. Besides, we conduct ablation studies to demonstrate the separate effect of each QIC module.
\end{itemize}
\section{Related Work}
\subsubsection{Medical Image Segmentation}
Deep learning with CNNs has become the dominated approach to MIS on various tissues, anatomical structures and lesions. Fully Convolutional Networks (FCNs) stand out to be a powerful deep net architecture for semantic segmentation~\cite{long2015fully}. FCNs replace the fully connected layers of standard CNNs with fully convolutional layers. Given an arbitrary-sized input image, FCNs can produce pixel-wise output of consistent size. Afterwards, the encoder-decoder networks become the major architecture for semantic segmentation, among which U-Net is well-recognized as a SOTA network for MIS~\cite{ronneberger2015}. U-Net adopts a symmetrical encoder-decoder architecture infused with skipped connections. In the downsampling path, it applies CNN, while in the upsampling path, it enhances the feature map, making its size be consistent with that of the input.

Following U-Net, several FCNs have been developed, such as U-Net 3D~\cite{cciccek2016}, V-Net~\cite{milletari2016}, Y-Net~\cite{mehta2018}, attention U-Net~\cite{oktay2018} and nnU-Net~\cite{isensee2018}, which can be regarded as different variants of U-Net. All these CNN type models require abundant expert-annotated data. When only a few labeled images of an ``unseen" class are available for segmenting an object of this class, these methods fail to provide good performance. When facing data-scarce cases, one needs to integrate data augmentation or transfer learning techniques to such methods to make them work.

\subsubsection{Few-shot Semantic Segmentation}
Different from supervised learning methods that require a large number of labeled samples to work, few-shot learning (FSL) methods aim to learn with few labeled samples. Thus, it is natural to adapt FSL methods to deal with MIS, leading to FSS methods. \cite{ouyang2020self} propose an adaptive local prototype pooling module to add extra focus on the background. \cite{hansen2022anomaly} use a single foreground prototype, and view image segmentation from an anomaly detection perspective. \cite{rakelly2018few} use weak annotations by replacing the pre-defined binary mask with a few selected landmarks of foreground and background. Most of these FSS methods aim to fully exploit information covered in the support set and assume that the labeled training data is abundant.

The prototypical network, which uses masked average pooling to extract class-wise features from the support set, is another major category of FSS methods featured by its strong interpretability and robustness against noises. \cite{dong2018few} adopt the meta-learning mechanism in prototypical networks for semantic segmentation. The resulting model contains a prototype network for learning class-specific prototypes and a segmentation network for making predictions on query images. \cite{wang2019panet} formulate image segmentation as an non-parametric prototype matching process, then propose prototype alignment network (PANet), which inversely predicts labels of the support images by using query images as the support set. \cite{liu2020part} improve PANet by introducing additional prototypes to capture more diverse features of the semantic classes. Base on PANet, \cite{ouyang2020self} and \cite{hansen2022anomaly} employ superpixel/supervoxel segmentation for a self-supervised prototypical FSS on medical images. \cite{tang2021recurrent} propose a context relation encoder and mask refinement module to iteratively refine the segmentation.

Another branch of medical FSS methods follows a two-arm architecture of \cite{shaban2017one}, which consists of a conditioner arm and a segmenter arm, corresponding to the support set and the query data, respectively. \cite{roy2020squeeze} employ dense connections with squeeze and excitation blocks to strengthen the interaction between the conditioner arm and the segmentor arm. \cite{sun2022few} present a global correlation network with discriminative embedding, obtaining an improved performance.

The aforementioned works do not explicitly take into account possible distribution shifts between the query image and the support set, while such shifts indeed exist in some practical MIS scenarios. Our proposed Q-Net learns to borrow information from the query image based on a meta-learning setting to remove or at least mitigate such distribution shifts.
\section{Problem Statement}
We consider an image segmentation task. First, we learn a segmentation model based on a training set $D_{train}$ with a label set $C_{train}$. Then we evaluate the model using a test set $D_{test}$, whose label set is denoted by $C_{test}$. For FSS tasks of our concern, $C_{train}$ and $C_{test}$ are disjoint, namely $C_{train}\cap C_{test}=\emptyset$.

We consider a meta-learning setting, in which both the training set $D_{train} = \{(S_i,Q_i)\}_{i=1}^{N_{train}}$ and the test set $D_{test} = \{(S_i,Q_i)\}_{i=1}^{N_{test}}$ consist of several randomly sampled episodes, where $N_{train}$ and $N_{test}$ are the number of episodes for training and testing, respectively. Each episode includes $K$ support images with annotations and a set of $Q$ query images of $N$ classes. That says we consider a N-way K-shot task. The support set $S_i = \{(\textbf{x}_k^s, \textbf{m}_k^s(c_j))\}_{k=1}^K$ contains $K$ image-mask pairs of a gray-scale image $\textbf{x} \in \mathbb{R}^{H\times W\times 1}$ and its corresponding binary mask $m \in \{0, 1\}^{H\times W}$ for class $c_j \in C_{train}, j=1,2,\dots, N$. The query set $Q_i$ contains $N_{qry}$ image-mask pairs from the same class as the support set. We learn the model on the support set and make a prediction for the query masks. Following the common practice in~\cite{ouyang2020self,hansen2022anomaly}, we adopt 1-way 1-shot meta-learning strategy for few-shot MIS.
\subsection{Supervoxel-guided Clustering}
Following \cite{hansen2022anomaly}, we use supervoxel clustering to generate pseudo labels for training, where each supervoxel produces a pseudo label. In this way, we can better utilize the volumetric nature of the medical images, compared with 2D superpixel clustering. We then construct episodic tasks for meta-training of our deep net model based on segmentations given by the supervoxels. Specifically, for each pseudo class/supervoxel, we randomly select one 2D image slice as a support image, and then use another adjacent image slice in the same volume as the query image. In what follows, we elaborate the proposed Q-Net, which is trained over such episodic tasks.
\section{Methodology}\label{sec:method}
\begin{figure*}[t]
\centering
\includegraphics[width=1.0\textwidth]{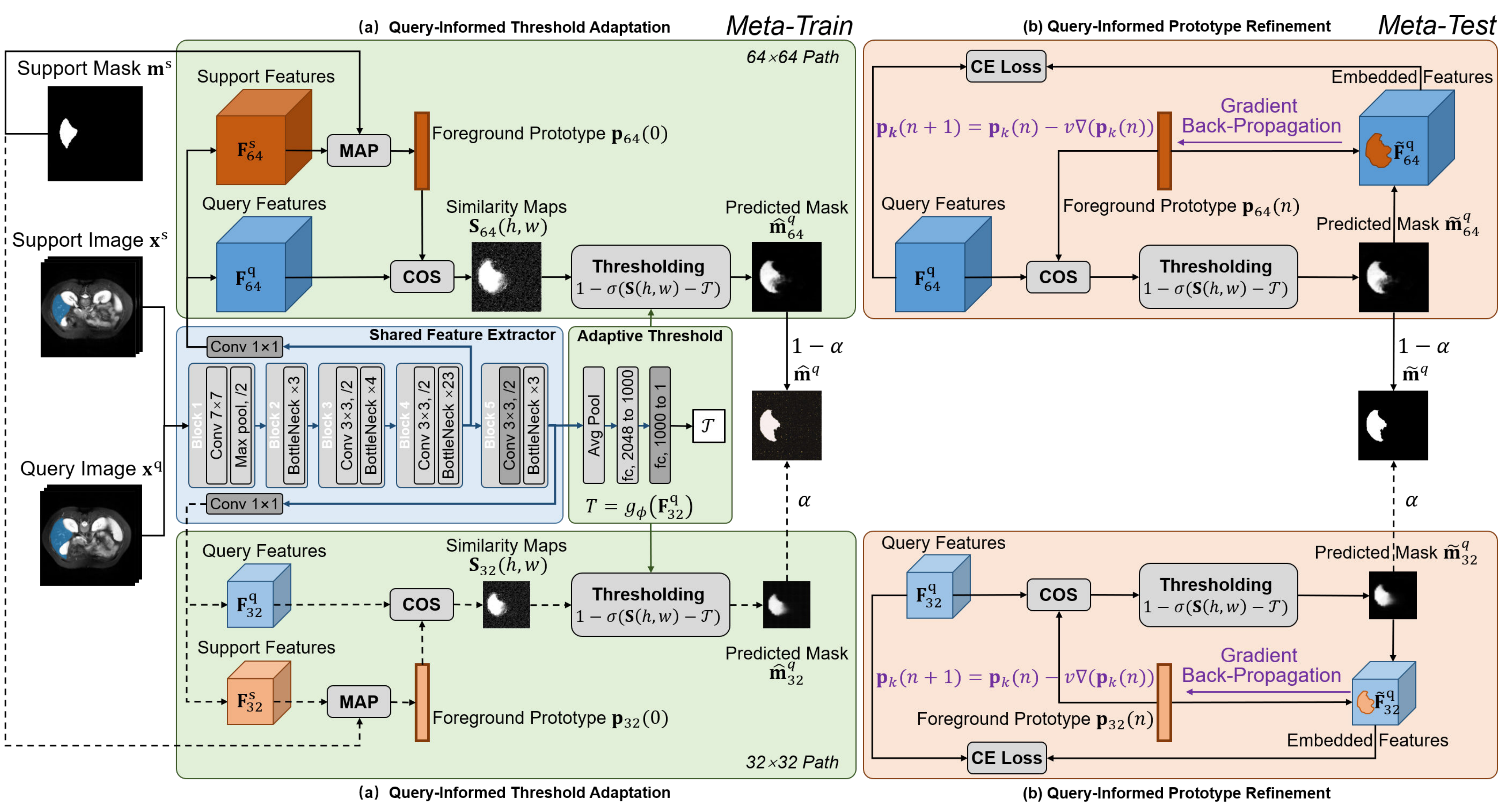}
\caption{\small Illustration of Q-Net. We use a shared feature encoder to learn deep feature maps in dual-path, corresponding to two feature scales: $32\times32$ and $64\times64$. The flows of operations are the same for these two paths, while we modularize the operations of the 2nd path and drawn them in gray to save space. In each path, we first learn one foreground prototype $\textbf{p}$ from the support features. Next we compute the similarity map between each query feature vector and the prototype. Then we predict the initial segmentation mask $\tilde{\textbf{m}}^q_{32}$ via anomaly detection performed on the similarity map with threshold $T$. See the Subsection \ref{sec:ta} for our approach to learn an adaptive $T$ value.
Then we refine the prototype by repeating the following two operations for a fixed number of times: (1) replacing the foreground feature vectors with the prototype; (2) minimizing a reconstruction loss. See more details in Section \ref{sec:method}.}
\label{fig2}
\end{figure*}

The proposed Q-Net is performed in three steps in test time: 1) extracting dual-path image features for both the support and the query images; 2) computing a query-informed anomaly threshold and then using it to separate the foreground object from the background; 3) refining the initial class prototype by incorporating query information and then using the refined prototype to make final segmentation. Figure~\ref{fig1} gives a brief conceptual illustration of our method, and Figure~\ref{fig2} shows a more detailed working flow of our method in test-time. Note that the model parameters are determined during meta-training, where the flow of the model stay unchanged but the prototype refinement module turns off.
\subsection{Dual-Path Feature Extraction}\label{sec:dp}
We employ a shared feature extractor $f_{\theta}$ to extract double-scale features from both the support and the query images in the embedding space. We denote the support features as $\textbf{F}^s=f_\theta(\textbf{x}^s), \textbf{F}^s\in\mathbb{R}^{H'\times W'\times Z}$, where $H'$, $W'$ and $Z$ denote the height, width and channel depth of the feature map. The query features $\textbf{F}^q=f_\theta(\textbf{x}^q)$ is defined in the same way.

Following the common practice, we use ResNet-101 with pre-trained MS-COCO weights as the backbone net for feature extraction. It is composed of a $7\times 7$ convolutional layer with stride of 2, a max pooling layer, and four residual blocks with multiple 3-layer bottleneck sub-blocks. The output features of the first residual block in the network are 4 times smaller than the input resolution. The network can provide another 3 times down-sampling in the last 3 residual blocks with dilation. In contrast to previous works~\cite{wang2019panet,ouyang2020self,hansen2022anomaly}, where their feature extractors output features that are 8 times smaller than input resolution by using dilation in the second residual block, we use dilation in the last residual block and extract features with two different scales from the last two residual blocks. This results in the resolution of double-scale features being 1/4 and 1/8 of the image resolution $(H'=H/4\ or\ H/8,\ W'=W/4\ or\ W/8)$. The double-scale feature extraction can better capture class prototypes of tissues of different sizes and shapes. The details of the shared feature extractor are illustrated in Figure~\ref{fig2}.
\subsection{Query-Informed Threshold Adaptation}\label{sec:ta}
Following~\cite{hansen2022anomaly}, we consider a single foreground prototype to avoid introducing additional background prototypes in each training episode. The prototypical prediction is undertaken by non-parametric metric learning. First, we compute the foreground prototype $\textbf{p} \in \mathbb{R}^{1\times Z}$ from the support feature map $\textbf{F}^s$ via masked average pooling:
\begin{small}
\begin{equation}
    \mathbf{p} = \frac{\sum_{i,j} \mathbf{F}^s(i,j) \odot \mathbf{m}^s(i,j)}{\sum_{i,j} \mathbf{m}^s(i,j)}  \\
\end{equation}
\end{small}
where the support feature map $\textbf{F}^s$ is resized to the mask size $(H,W)$, and $\odot$ denotes the Hadamard product for masked average pooling.

In order to perform soft thresholding on the predicted mask, we employ a shifted Sigmoid on the negative cosine similarity \textbf{S}(h,w) between the query feature vectors and the foreground prototype \textbf{p}, to generate a probability map of foreground $\hat{\textbf{m}}^q_f$, as shown in Equations (\ref{eq:neg_cosine})-(\ref{eq:thresh}).
\begin{small}
\begin{equation}
    \textbf{S}(h,w) = -a \frac{\mathbf{F}^q(h,w) \cdot \textbf{p}}{\Vert \mathbf{F}^q \Vert \Vert \textbf{p} \Vert} \label{eq:neg_cosine}
\end{equation}
\begin{equation}
    \hat{\textbf{m}}^q_f = 1 - \sigma(\textbf{S}(h,w) - T) \label{eq:thresh}
\end{equation}
\begin{equation}
    \hat{\textbf{m}}^q_b = 1 - \hat{\textbf{m}}^q_f
\end{equation}
\end{small}
where $(h,w)$ denotes the spatial locations of the foreground mask, $a=20$ a scaling factor~\cite{oreshkin2018tadam}, and $T$ the learned threshold. The subscripts $_f$ and $_b$ denote ``foreground" and ``background", respectively.

In common practice of FSS, when $T$ is learned in the meta-training stage, it is then directly used in the meta-testing stage to segment unseen foreground objects. In the standard setting, the background may contain novel test objects, then the supervoxels overlapped with the test labels might be used as the training classes. Compared to the spatially heterogeneous background, the foreground objects, such as the right and the left kidneys in the abdominal MRI image, are homogeneous, thus can be easily clustered into the same supervoxel. The learned threshold used for anomaly detection has been optimized on supervoxels that share similar shape and size to the ``unseen" objects. For such cases, using a fixed $T$ makes sense. However, if image slices containing the test classes are removed, then the test classes becomes completely unseen classes for the algorithm agent, then the threshold learned in the training process can not be adapted straightforward to such ``unseen" classes.

Following the QIL strategy, we add two fully-connected layers $g_{\phi}$ at the top of the feature extractor $f_{\theta}$ to learn a soft threshold from the query image instead of the predicted foreground masks on the seen classes:
\begin{small}
\begin{equation}
    T = g_{\phi}(\textbf{F}^q)
\end{equation}
\end{small}
where the fully-connected layer maps the number of channels of the flattened feature maps from 2048 to 1.  The two predicted masks in our dual-path network are then up-sampled to $(H,W)$ and further combined through element-wise summation with a balance factor $\alpha \in (0,1.0)$ as follows.
\begin{equation}
    \hat{\mathbf{m}}^q = \alpha\cdot \hat{\mathbf{m}}_{64}^q + (1-\alpha)\cdot \hat{\mathbf{m}}_{32}^q\label{eq:alpha}
\end{equation}
See more details in the Appendix subsection \label{sec:appendix1}.
\subsection{Query-Informed Prototype Refinement}\label{sec:pr}
Given class prototypes extracted from the support images, we use cosine similarity to make segmentation predictions on the foreground and background. This prediction process assumes that a query image holds the similar appearance with support images. During the training phase, the image slices adjacent to the support image are selected as query samples, while during the inference phase, the query image and the support set can be selected from different image volumes. Therefore, a query image can be quite different from the support images on intensity and appearance. Following the QIL strategy, We propose a prototype refinement module to update the class prototypes by borrowing information from the query image features in test time.

In prototype refinement, firstly we produce the predicted mask by calculating the cosine similarity between the query image features and the class prototypes. Given the foreground prototype $\mathbf{p}(n)$ and the predicted mask $\tilde{\mathbf{m}}^q(n)$ in each scale at the $n$-th iteration, we apply element-wise multiplication between the query feature maps $\mathbf{F}^q$ and $\tilde{\mathbf{m}}^q(n)$ to obtain its background feature vectors. Then, we replace its foreground feature vectors in $\mathbf{F}^q$ with $\mathbf{p}(n)$ to generate new query feature maps $\tilde{\mathbf{F}}^q$ as follows
\begin{small}
\begin{equation}
    \tilde{\mathbf{m}}^q(n)(i,j) = \left\{
    \begin{array}{rcl}
    1, & & \mbox{if}\quad \tilde{\mathbf{m}}^q_b(n)(i,j) \leq \tilde{\mathbf{m}}^q_f(n)(i,j)\\[1ex]
    0, & & \mbox{else}
    \end{array}
    \right.
\end{equation}
\begin{equation}
    \tilde{\mathbf{F}}^q(i,j) = \left\{
    \begin{array}{rcl}
    \mathbf{F}^q(i,j), & &\mbox{if}\quad \tilde{\mathbf{m}}^q(n)(i,j) = 1\\
    \mathbf{p}(n), & &\mbox{if}\quad  \tilde{\mathbf{m}}^q(n)(i,j) = 0
    \end{array}
    \right.
\end{equation}
\end{small}
We design this module to mimic the steps of an optimization algorithm in inference time. This module is trained to modify the foreground prototypes $\textbf{p}$ gradually so that the final predicted mask $\hat{\textbf{m}}^q$ converges to an optimum solution.
\begin{small}
\begin{equation}
    \mathbf{p}_k(n+1) = \mathbf{p}_k(n) - v\partial\sum_{n=1}^N \mathcal{L}(\tilde{\mathbf{F}}_k^q, \mathbf{F}_k^q)/\partial \mathbf{p}_k(n)
\end{equation}
\end{small}
where $\mathcal{L}(\tilde{\mathbf{F}}^q_k, \mathbf{F}^q_k)$ is the cross-entropy loss between the new query features $\tilde{\mathbf{F}}^q_k$ and the original query features $\mathbf{F}^q_k$, and $N$ denotes the number of iterations of gradient back-propagation and $i$ denotes 32 or 64 corresponding to the double-scale paths. $\textbf{p}_N$ denotes the optimal prototype after $N$ iterations. For more implementation details on prototype refinement, readers are referred to the Appendix subsection \label{sec:appendix2}.
\subsection{Loss function}\label{sec:loss}
We train our network by minimizing the cross-entropy between the predicted masks and the ground-truth segmentation $\textbf{m}^q$. The dual-path predicted masks after anomaly threshold adaptation are up-sampled to $(H,W)$ and then combined through element-wise summation, as follows
\begin{small}
\begin{equation}
    \hat{\textbf{m}}^q = \alpha\cdot \hat{\textbf{m}}_{64}^q + (1-\alpha)\cdot \hat{\textbf{m}}_{32}^q
\end{equation}
\end{small}
where $\alpha$ is a balance factor, whose value is set empirically, see Table \ref{table3} for an ablation analysis.

We compute the binary cross-entropy loss as the segmentation loss for each training episode:
\begin{small}
\begin{equation}
    L_{seg} = -\frac{1}{HW}\sum_h^H \sum_w^W \sum_{j=\{f,b\}} \textbf{m}_j^q(h,w) \log(\hat{\textbf{m}}_j^q(h,w))
\end{equation}
\end{small}
Following~\cite{wang2019panet,ouyang2020self}, we inversely predict labels of the support images by using query images as the support set, then build a prototypical alignment regularization item as follows
\begin{small}
\begin{equation}
    L_{reg} = -\frac{1}{HW}\sum_h^H \sum_w^W \sum_{j=\{f,b\}} \textbf{m}_j^s(h,w) \log(\hat{\textbf{m}}_j^s(h,w))
\end{equation}
\end{small}
Overall, the loss function for each training episode is defined to be
\begin{small}
\begin{equation}
    L = L_{seg} + L_{reg}
\end{equation}
\end{small}

For prototype refinement, we compute pixel-wise binary cross-entropy loss between the embedded feature maps $\tilde{\textbf{F}}^q_i$ and the original query feature maps $\textbf{F}^q_i$ to update class prototype $\textbf{p}_i$. These feature maps are firstly normalized to [0,1], and then put into the pixel-wise binary cross-entropy loss function.
\section{Experiments}
\subsection{Datasets}
We evaluate the proposed Q-Net on two popular MRI datasets for FSS, i.e., \textbf{ABD} \cite{kavur2021chaos} and \textbf{CMR}~\cite{zhuang2016multivariate}.

\textbf{ABD} is an abdominal MRI dataset published from the ISBI 2019 Combined Healthy Abdominal Organ Segmentation Challenge (CHAOS). It includes 20 3D T2-SPIR MRI scans with on average 36 slices from \textit{liver, left kidney, right kidney}, and \textit{spleen}.

\textbf{CMR} is a MRI dataset published from the MICCAI 2019 Multi-sequence Cardiac MRI Segmentation Challenge (bSSFP fold). It contains 35 3D cardiac MRI scans with on average 13 slices.

\subsection{Performance Metric}
Following the common practice in medical few-shot image segmentation, we use the mean S\o rensen-Dice coefficient (DSC) as evaluation metric. It measures the overlap ratio of the predicted mask $\V{A}$ and the ground-truth segmentation $\V{B}$ as follows:
\begin{small}
\begin{equation}
    DSC(\V{A}, \V{B}) = \frac{2\Vert\V{A} \cap \V{B}\Vert}{\Vert\V{A} \Vert+\Vert\V{B} \Vert} * 100\%
\end{equation}
\end{small}
The greater this DSC score, the better the segmentation performance, and vice versa.

\subsection{Data Pre-processing}
In our experiments, we adopt the same image pre-processing scheme as in~\cite{ouyang2020self}, to make a fair performance comparison. Specifically, We first cut off the bright end (the top 0.5\%) of the histogram to alleviate the off-resonance issue. Then we re-sample the image slices to get the same resolution as before. We use axial slices for ABD and short-axis slices for CMR, respectively. At last, we crop these slices to an unified size of $256 \times 256$ pixels.

We implement the proposed method using the PyTorch (v1.10.2) on a NVIDIA RTX 3090Ti GPU. The SGD optimizer is employed to update the network weights with configuration parameters of $\beta_1=0.9,\beta_2=0.999$, and $\varepsilon=10^{-8}$.
In the meta-training phase, the learning rate is initially set to 0.001 and decayed by a rate of 0.98. The training of the model is conducted for up to 30k iterations.

\subsection{Experimental Settings}
We consider two experimental settings to evaluate the performance of our model.

\textbf{Setting 1} is the standard setting, where objects of test classes can appear in the background in the training dataset. Since we perform a self-supervised supervoxel-based segmentation on the whole image, the resulting supervoxels have similar shapes and sizes as objects of the test classes. Therefore, objects of test classes may be implicitly involved in the training process, then the test classes are not truly ``unseen" classes for the algorithm agent, in this setting.

\textbf{Setting 2} generalizes the Setting 1, for which we directly remove image slices that contains the test classes from the training data. It guarantees that the test classes are truly ``unseen" classes for the model. We follow the same protocol in~\cite{ouyang2020self} to separate the four testing organs into two groups: upper abdomen of liver and spleen, and lower abdomen of left/right kidneys.

\subsection{Results}
\subsubsection{Comparison with SOTA methods}
We compare Q-Net with modern SOTA models, including SE-Net~\cite{roy2020squeeze}, PANet~\cite{wang2019panet}, ALPNet~\cite{ouyang2020self}, and ADNet~\cite{hansen2022anomaly}, for both settings mentioned above. See the result in Table~\ref{table1}. \cite{ouyang2020self} reported performance for SE-Net on ABD and CMR, so these results are directly quoted. We ran algorithms of PANet, ALPNet and ADNet using public available code to report their performance on ABD and CMR. It shows that our Q-Net significantly outperforms the SOTA methods on both datasets.

Specifically, for setting 1 where the test organs may appear in the background, Q-Net gives the largest mean DSC score on both ABD and CMR. In particular, its dice score for right-kidney on ABD achieves about 88\%. For setting 2 where the test organs are completely removed from the background, Q-Net performs best again, while the SOTA ADNet performs poorly for small-sized organs, such as the right kidney. This result indicates that Q-Net is better at segmenting small-sized organs, compared with ADNet.

For dataset CMR, we only consider Setting 1, as Setting 2 is impractical for cardiac MRI scans. The left-ventricle myocardium (LV-MYO) is wrapped in the left ventricle blood pool (LV-BP) and the right-ventricle (RV) is quite close to them. We see that Q-Net performs much better than the others in segmenting such adjacent organs, especially for LV-BP and RV. Although foreground objects in the CMR images have smaller variations in size, Q-Net still performs best as shown in bottom lines of Table \ref{table1}.
\begin{table*}[!htp]
\caption{\small DSC comparison with other methods on ABD and CMR datasets. Bold and italic numbers denote the best and second best results, respectively.}
\small
\centering
\begin{tabular}{|c|l|c|c|c|c|c|}
  \hline
  \multicolumn{7}{|c|}{\textbf{ABD}}\\
  \hline
  \textbf{Settings} & \textbf{Method} & \textbf{Liver} & \textbf{R.kidney} & \textbf{L.kidney} & \textbf{Spleen} & \textbf{mean} \\
  \hline
  \multirow{5}{*}{Setting 1} & SE-Net & 29.02 & 47.96 & 45.78 & 47.30 & 42.51 \\
  & PANet & $73.81\pm3.39$ & $80.67\pm3.35$ & $69.48\pm6.05$ & $69.10\pm10.18$ & $73.27\pm7.90$ \\
  & ALPNet & $78.55\pm2.48$ & $83.11\pm5.35$ & \textit{78.16$\pm$6.54} & $70.58\pm6.16$ & $77.60\pm7.01$ \\
  & ADNet & $\mathbf{82.11\pm2.19}$ & \textit{85.80$\pm$4.51} & $73.86\pm8.23$ & \textit{72.29$\pm$8.87} & \textit{78.51$\pm$8.63} \\
  & Q-Net & \textit{81.74$\pm$3.83} & $\mathbf{87.98\pm4.55}$ & $\mathbf{78.36\pm8.36}$ & $\mathbf{75.99\pm8.64}$ & $\mathbf{81.02\pm8.08}$ \\
  \hline
  \multirow{5}{*}{Setting 2} & SE-Net & 27.43 & 61.32 & 62.11 & 51.80 & 50.66 \\
  & PANet & $69.37\pm5.44$ & \textit{66.94$\pm$5.67} & \textit{63.17$\pm$7.84} & $61.25\pm7.84$ & $65.68\pm7.54$ \\
  & ALPNet & $70.73\pm3.81$ & $\mathbf{73.30\pm9.72}$ & $61.20\pm8.22$ & \textit{62.98$\pm$9.28} & \textit{67.05$\pm$9.57} \\
  & ADNet & \textit{77.03$\pm$3.36} & $56.68\pm10.84$ & $59.64\pm9.04$ & $59.44\pm7.46$ & $63.20\pm11.48$ \\
  & Q-Net & $\mathbf{78.25\pm4.81}$ & $65.94\pm10.08$ & $\mathbf{64.81\pm7.42}$ & $\mathbf{65.37\pm8.61}$ & $\mathbf{68.59\pm9.73}$ \\
  \hline
  \hline
  \multicolumn{7}{|c|}{\textbf{CMR}}\\
  \hline
  \textbf{Settings} & \textbf{Method} & \textbf{LV-BP} & \textbf{LV-MYO} & \textbf{RV} & \multicolumn{2}{|c|}{\textbf{mean}} \\
  \hline
  \multirow{5}{*}{Setting 1} & SE-Net & 58.04 & 25.18 & 12.86 & \multicolumn{2}{|c|}{32.02} \\
  & PANet & $72.77\pm7.66$ & $44.76\pm4.08$ & $57.13\pm4.54$ & \multicolumn{2}{|c|}{$58.20\pm12.79$} \\
  & ALPNet & $85.42\pm3.95$ & $63.38\pm3.95$ & $74.07\pm4.47$ & \multicolumn{2}{|c|}{$74.29\pm9.90$} \\
  & ADNet & \textit{86.26$\pm$2.18} & \textit{65.08$\pm$4.85} & \textit{76.50$\pm$1.97} & \multicolumn{2}{|c|}{\textit{75.95$\pm$9.34}} \\
  & Q-Net & $\mathbf{90.25\pm1.44}$ & $\mathbf{65.92\pm2.96}$ & $\mathbf{78.19\pm2.95}$ & \multicolumn{2}{|c|}{$\mathbf{78.15\pm10.22}$} \\
  \hline
  \end{tabular}
  \label{table1}
\end{table*}
\begin{table*}[t]
\caption{\small Ablation study for Q-Net on dataset ABD under Setting 2. 'DS' means downsampling. 'DP', 'TA' and 'PR' denote dual-path feature extraction, query-informed threshold adaptation, and query-informed prototype refinement, respectively.}
\centering
\small
\begin{tabular}{|l|l|c|c|c|c|c|}
  \hline
  \textbf{Experiment} & \textbf{Method} & \textbf{Liver} & \textbf{R.kidney} & \textbf{L.kidney} & \textbf{Spleen} & \textbf{mean} \\
  \hline
  \multirow{3}{*}{Feature Extractor} & DP (DS on Layer 2) & $72.56$ & $55.91$ & $52.61$ & $60.65$ & $60.43\pm15.90$ \\
  & DP (DS on Layer 3) & $71.00$ & $56.74$ & $55.05$ & $66.01$ & $62.20\pm12.02$ \\
  & \textbf{DP (DS on Layer 4)} & $77.59$ & $62.08$ & $62.86$ & $62.42$ & $66.24\pm9.62$ \\
  \hline
  \multirow{6}{*}{Added components} & $32\times32$ & $77.03$ & $56.68$ & $59.54$ & $59.44$ & $63.20\pm11.48$ \\
  & $64\times64$ & $78.65$ & $59.58$ & $60.82$ & $62.80$ & $65.46\pm11.40$ \\
  & DP & $78.53$ & $61.35$ & $64.30$ & $62.55$ & $66.68\pm9.99$ \\
  & DP + PR & $78.35$ & $62.91$ & $64.94$ & $63.13$ & $67.33\pm9.65$ \\
  & DP + TA & $78.79$ & $63.87$ & $64.69$ & $64.38$ & $67.93\pm10.09$ \\
  & \textbf{DP + TA + PR} & $78.25$ & $65.94$ & $64.81$ & $65.37$ & $68.59\pm9.73$ \\
  \hline
  \end{tabular}
  \label{table2}
\end{table*}

\begin{figure}[t]
	\centering
	\includegraphics[width=\linewidth,scale=0.95]{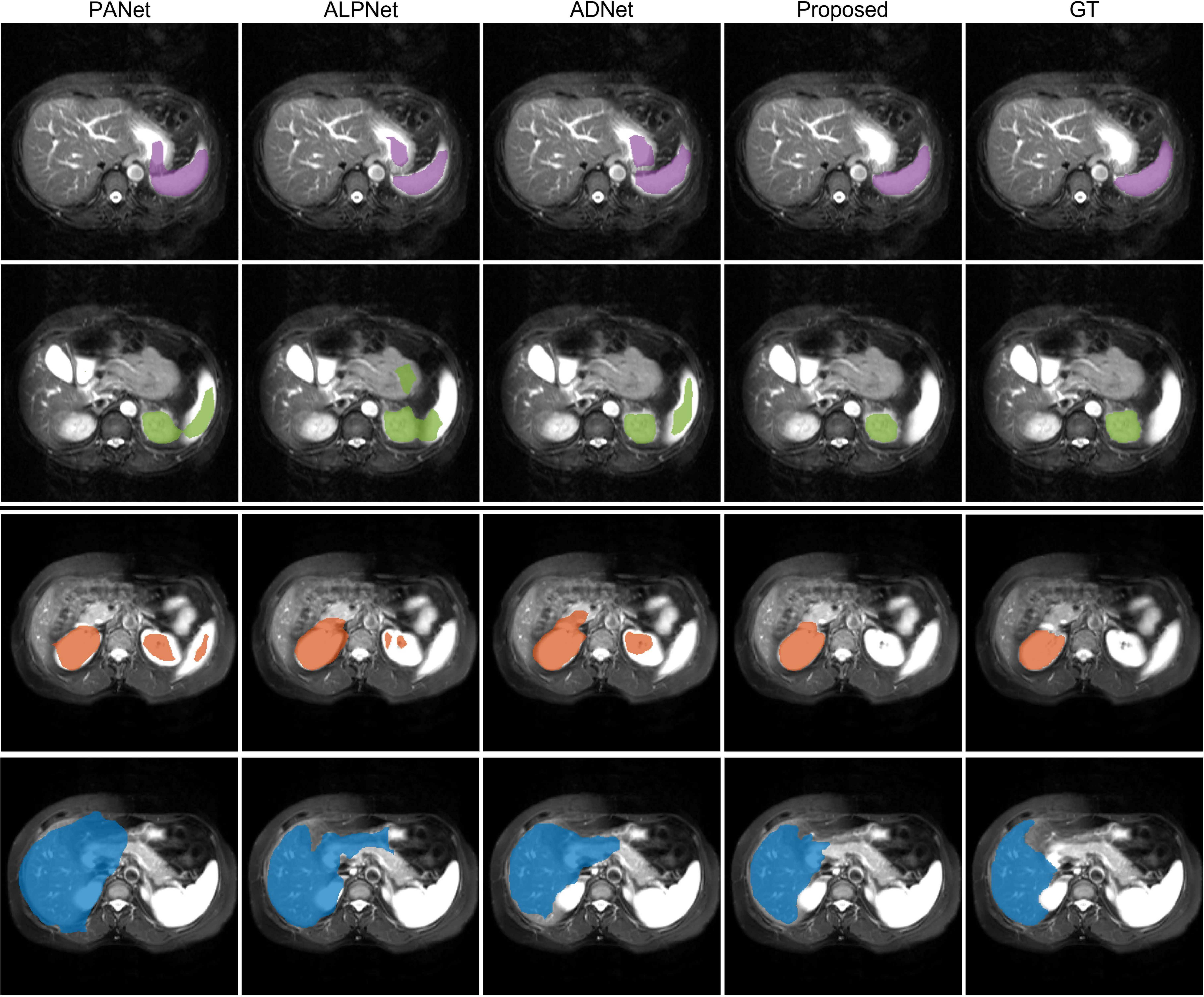}
	\caption{\small  Qualitative comparisons for the ABD dataset. Left to right: Segmentation results and ground-truth segmentation of a query slice containing the target object. Top to bottom: spleen, left kidney, right kidney and liver. (Best viewed with zoom)}
	\label{fig3}
\end{figure}

We also provide some qualitative comparisons in Figure~\ref{fig3}. It is shown that Q-Net is more robust against variations of the objects' appearance patterns. It also reconfirms that Q-Net performs better in segmenting small-sized organs.

\subsubsection{Ablation Study}
First, we do an ablation study for Q-Net, to test how the way in which we construct the $64\times 64$ feature maps affects the final segmentation performance. In the ResNet backbone, there are three residual blocks containing the dilation operation. We adopt the third downsampling operation by activating its dilation operation in one of these residual blocks (i.e., layers 2, 3 and 4). The outputs of the last residual block before the third downsampling component are $64\times 64$ feature maps. The segmentation results on ABD dataset under Setting 2 are shown in Table~\ref{table2}, where DS denotes downsampling. We observe that using the downsampling component in the last residual block can provide a better combination of double-scale predictions.

Second, we test separate contributions of our three proposed computation modules, namely query-informed threshold adaptation (TA), query-informed prototype refinement (PR), and the dual-path extension (DP), by conducting an ablation study of them on the ABD dataset, under Setting 2. The result is presented in Table~\ref{table2}. We see from the last column that each module indeed makes a separate contribution to improve the final segmentation performance. In particular, DP brings a performance gain of 1.22\%, and the two query-informed modules contribute an additional gain of 1.91\%.

In the third study, we evaluate the effect of the balance factor $\alpha$ in Equation (8) on dual-path predictions. $\alpha$ denotes the weight of the larger feature maps' contribution in producing the final mask prediction. As shown in Table~\ref{table3}, For Setting 1, the larger the $\alpha$, the better, while for Setting 2, the optimal $\alpha$ value is 0.8.
\begin{table}[H]
\caption{\small Ablation study for the effect of $\alpha$ on the performance of Q-Net in terms of the DSC score.}
\centering
\small
\begin{tabular}{|c|c|c|}
  \hline
  \multirow{2}{*}{$\mathbf{\alpha}$} & \multicolumn{2}{|c|}{ABD}\\
  \cline{2-3}
  & Setting1 & Setting2\\
  \hline
  0.9 & $81.02\pm8.08$ & $67.28\pm11.68$ \\
  0.8 & $80.31\pm8.45$ & $68.75\pm11.71$ \\
  0.6 & $80.28\pm8.33$ & $68.59\pm9.73$ \\
  0.5 & $79.72\pm9.00$ & $67.98\pm10.47$ \\
  0.4 & $79.30\pm9.11$ & $66.89\pm10.09$ \\
  0.2 & $78.64\pm8.74$ & $65.89\pm10.49$ \\
  \hline
  \end{tabular}
  \label{table3}
\end{table}

\begin{figure}[H]
	\centering
	\includegraphics[width=0.7\linewidth]{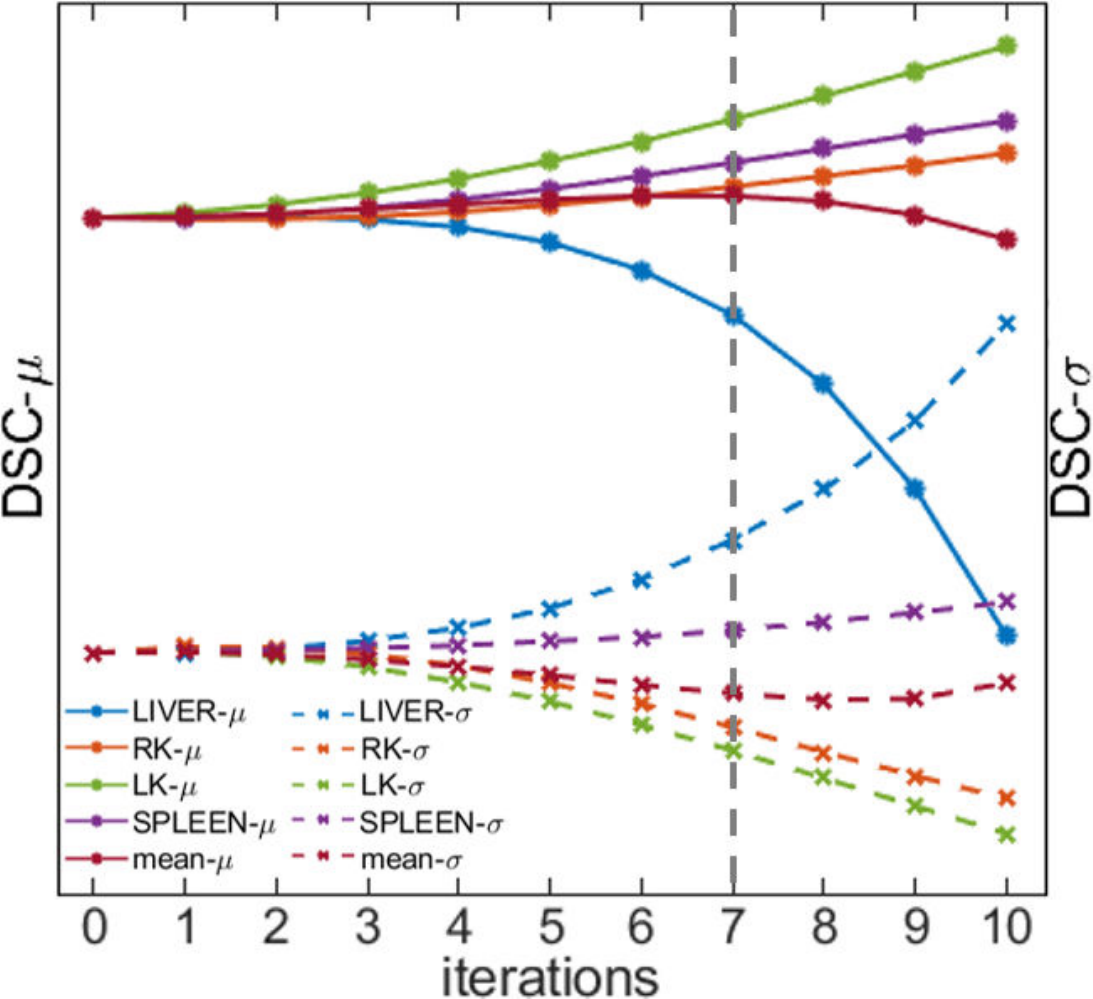}
	\caption{\small DSC scores collected at different checkpoints of the iterative prototype refinement process. $\mu$ and $\sigma$ denote the mean and the standard error, respectively.}
	\label{fig4}
\end{figure}

Finally, we test the effect of the number of iterations for prototype refinement on the final performance of Q-Net. We do this experiment on dataset ABD with a learning rate of 0.01, and compute DSC scores corresponding to different checkpoints at the prototype refinement process. The result is shown in Figure~\ref{fig4}. We let the curves corresponding to different organs share the same starting point for ease of comparison. Figure~\ref{fig4} shows that there is a slight increase in DSC scores of left/right kidney and spleen, but a substantial fall in DSC score of liver. Such an analysis suggests an empirical choice of the iteration number at 7, where the mean DSC score over all four organs reaches the peak value.
\section{Discussions}
Currently Q-Net uses a single prototype for each organ, while this may be inappropriate for dealing with large organs with in-homogeneous structures, such as liver, which can occupy almost half of a medical image. Larger feature maps may exacerbate this problem. How to assign an adaptive number of prototypes for a specific organ deserves future research.

In addition, Q-Net treats each query image independently. That means the knowledge informed from the query images are not accumulated for future segmentation tasks. An future direction following this line is to investigate approaches to do incremental query-informed learning.

Finally, the query-informed adaptation (QIA) mechanism has a natural Bayesian flavor. A formal analysis on the tie between QIA and Bayesian is missing here and can be conducted in the future.
\section{Conclusions}
We proposed Q-Net, a novel meta-learning approach derived from ADNet to few-shot medical image segmentation. It is characterized by introducing two query-informed computational modules that enable dynamic threshold and prototype adjustments during inference for mitigating distribution shifts between the support set and the query image. Additionally, Q-Net employs a dual-path architecture to capture double-scale features. Experimental results on abdominal and cardiac magnetic resonance image datasets showed that Q-Net outperformed SOTA methods, especially for segmentation of small-sized organs. This work sheds light on how to improve meta-learning techniques for few-shot image segmentation by borrowing information from the query image.
\appendix
\section{Appendix}
\subsection{Implementation Details of Query-Informed Threshold Adaptation}\label{sec:appendix1}
\begin{figure}[htp]
	\centering
	\includegraphics[width=1.0\linewidth]{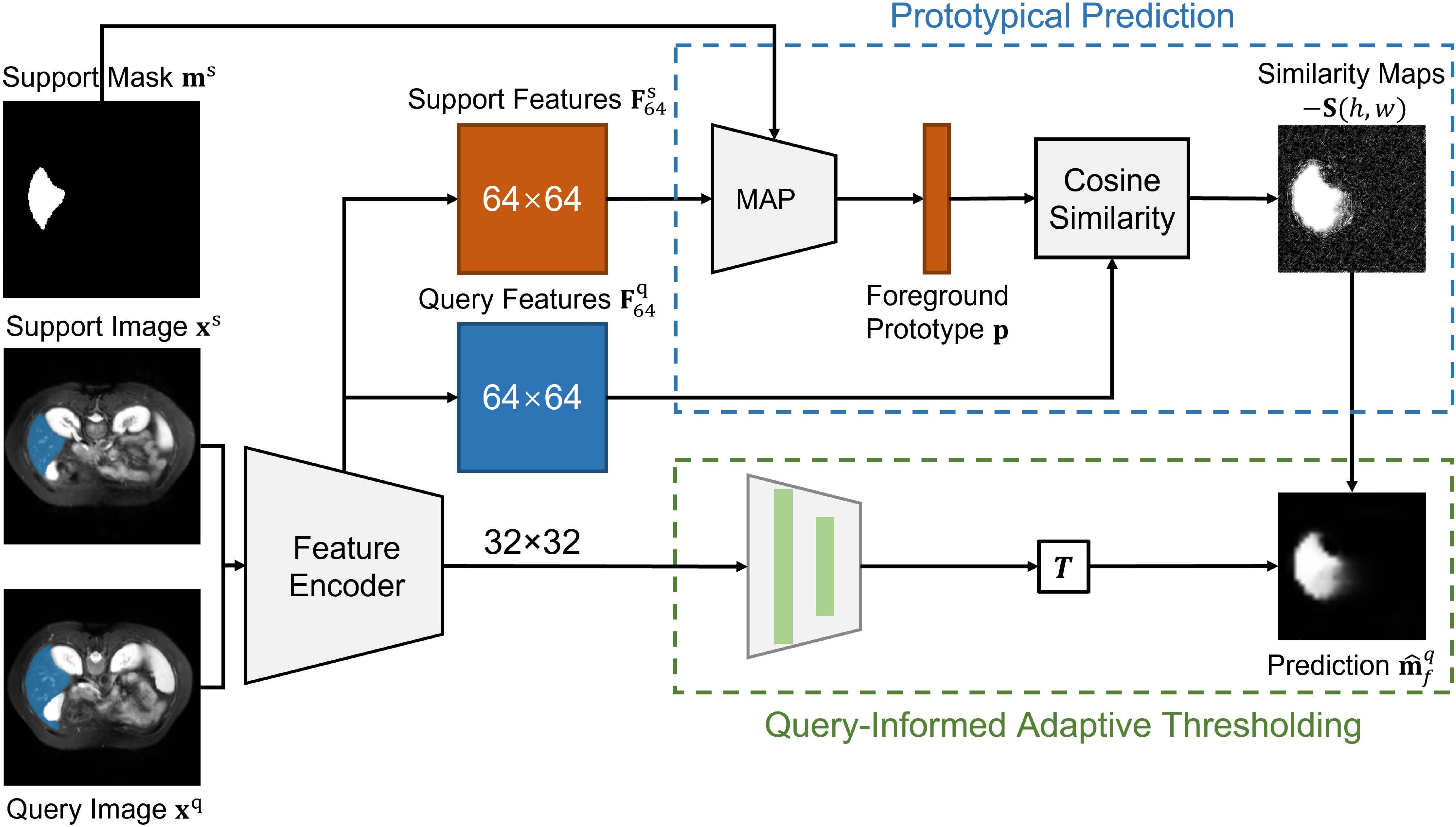}
	\caption{\small Illustration of the ``Query-informed Threshold Adaptation'' module.}
	\label{fig5}
\end{figure}

The computation module of Query-Informed Threshold Adaptation consists of two operations, namely prototype prediction and a threshold adaptation. We show the workflow of this module for the path corresponding to the feature scale of $64\times 64$ in Figure~\ref{fig5}.

Given the support feature $\textbf{F}^s_{64}$ and query feature $\textbf{F}^q_{64}$ with the size of $64\times 64\times 512$, we compute the foreground prototype $\textbf{p} \in \mathbb{R}^{1\times 512}$ from the support feature $\textbf{F}^s_{64}$ via masked average pooling. Then we compute the negative cosine similarity map $\textbf{S}(h,w)$ between each query feature vector and the foreground prototype, which has the same size as the support mask size $(H,W)$. This is the workflow of the prototypical prediction with only one foreground prototype.

In query-informed adaptive thresholding module, the adaptive threshold $T$ is learned from the query feature $\textbf{F}^q_{32}$ with the size of $32\times 32\times 512$, which is followed by an adaptive threshold generator $g_{\phi}$. The adaptive threshold generator includes two fully-connected layers that convert the channel size of the flattened feature maps from 2048 to 1000, then to one. This adaptive threshold $T$ is not a fixed value. It depends on the query image, which may contain an object of an ``unseen'' semantic class. In the stage of testing, parameters of the feature encoder and the adaptive threshold generator are fixed when implementing iterative prototype refinement.
\subsection{Implementation Details of Query-Informed Prototype Refinement}\label{sec:appendix2}
\begin{figure}[htp]
	\centering
	\includegraphics[width=1.1\linewidth]{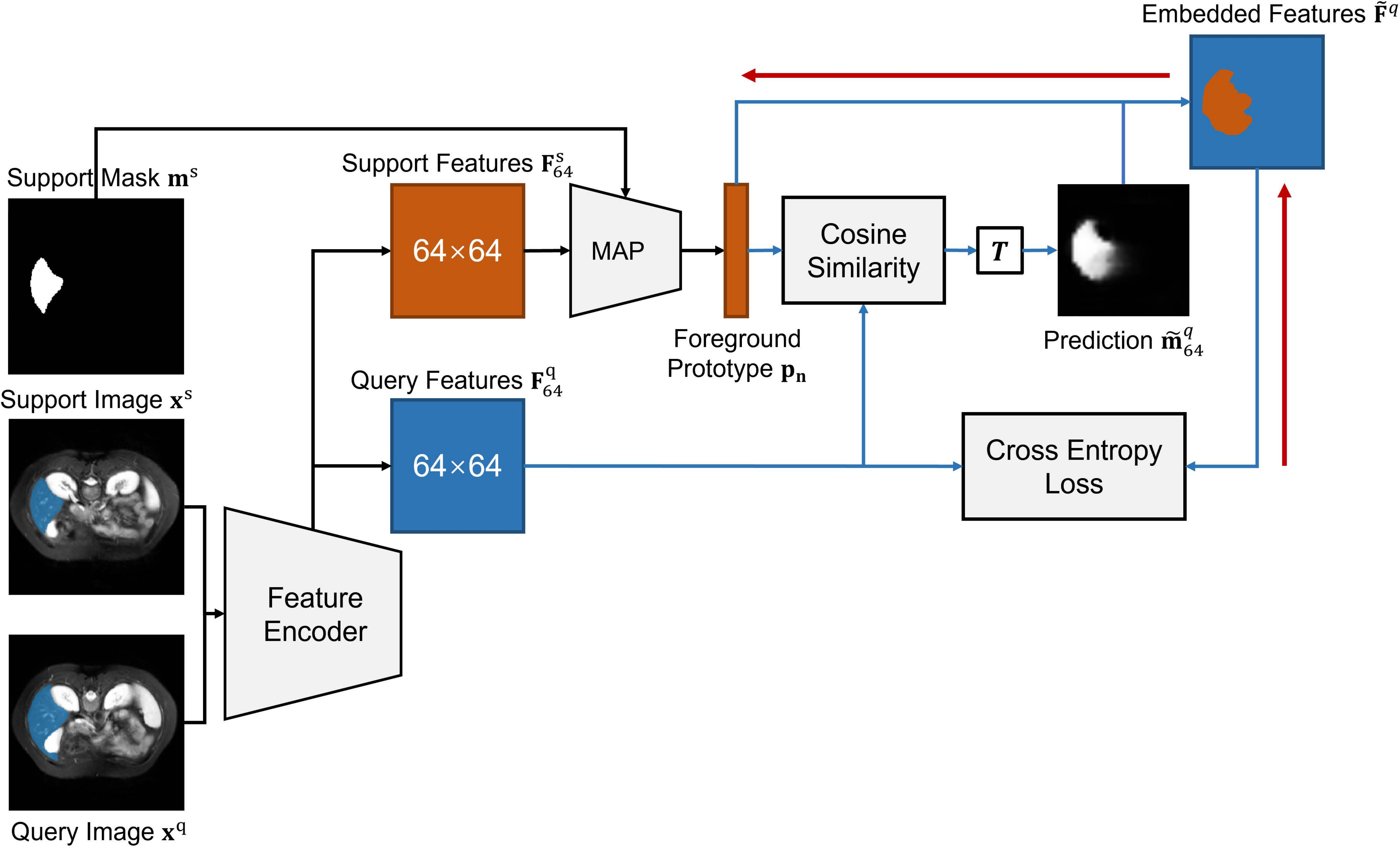}
	\caption{\small Illustration of the ``Query-informed Prototype Refinement'' module}
	\label{fig6}
\end{figure}

The Query-Informed Prototype Refinement module is only activated in the stage of testing. Figure~\ref{fig6} illustrates the iterative workflow of updating the foreground prototype guided by the gradient of loss, incurred when replacing the query feature vectors corresponding to the foreground part of the predicted mask by the single foreground prototype $\textbf{p}$. In Figure~\ref{fig6} the adaptive threshold $T$ is determined through the query-informed threshold adaptation module.

A pseudocode that implements the prototype refinement module corresponding to the feature scale of $64\times 64$ is presented in Algorithm~\ref{algorithm1}.
\begin{algorithm}[H]
\caption{Query-informed Prototype Refinement}
\label{algorithm1}
\textbf{Require}: Support feature $\textbf{F}^s_{64}$, support mask $\textbf{m}^s$, and query feature $\textbf{F}^q_{64}$\\
\begin{algorithmic}[1] 
\STATE Compute an initial foreground prototype $\textbf{p}_0$
\STATE Compute negative cosine similarity $\textbf{S}(h,w)$ between each query feature vector and
the foreground prototype. Set $n=0$.
\STATE Get a predicted mask $\tilde{\textbf{m}}^q_n$
\STATE Reallocate the foreground prototype to give a new query features $\hat{\textbf{F}}^q_n(i,j)$.

$\hat{\textbf{F}}^q_n(i,j) = \left\{
    \begin{array}{rcl}
    \textbf{F}^q(i,j), & & {\hat{\textbf{m}}^q_n(i,j) = 0}\nonumber\\
    \textbf{p}_n, & & {\hat{\textbf{m}}^q_n(i,j) = 1}
    \end{array}
    \right.$
\STATE Compute cross-entropy loss between the reconstructed query feature $\tilde{\textbf{F}}^q_n$ and the original query feature $\textbf{F}^q$ and update $\textbf{p}_n$

$\textbf{p}_{n+1} = \textbf{p}_n - v\frac{\partial\sum_{n=1}^N L(\hat{\textbf{F}}^q, \textbf{F}^q)}{\partial \textbf{p}_n}$
\STATE If the stopping criterion does not met, repeat from step 2, otherwise, output $\textbf{p}_{n+1}$.
\end{algorithmic}
\end{algorithm}
\bibliographystyle{IEEEbib}
\bibliography{Bibliography_File}
\end{document}